\documentclass[twoside,11pt]{article}

\usepackage[arxiv]{melba}

\usepackage{mwe} 

\usepackage{amsmath,amsfonts, booktabs}

\newcommand{\s}[1]{\left[ {#1} \right]}

\newcommand{\norm}[1]{\left|\left| {#1} \right|\right|}

\newcommand{\eqnref}[1]{Equation~\ref{#1}}

\melbaid{YYYY:NNN}  
\doi{https://doi.org/10.59275/j.melba.2023-AAAA}
\melbaauthors{Mahmood, Oliva, Styner}  
\volume{2}
\firstpageno{1}  
\melbayear{2024}  
\datesubmitted{06/2024}  
\datepublished{m2/yyyy}  

\ShortHeadings{Spatial-MSMA for Anomaly Detection}{Mahmood, Oliva and Styner}

\title{Localizing Anomalies via Multiscale Score Matching Analysis}

\author{\name Ahsan Mahmood \email amahmood@cs.unc.edu  
	\AND
	\name Junier Oliva  \email joliva@cs.unc.edu
        \AND
        \name Martin Styner  \email styner@cs.unc.edu \\
        \\
	\addr Department of Computer Science, University of North Carolina at Chapel Hill, Chapel Hill, NC, USA
}

\begin{document}

\maketitle

\begin{abstract}

 Anomaly detection and localization in medical imaging remain critical challenges in healthcare. This paper introduces Spatial-MSMA (Multiscale Score Matching Analysis), a novel unsupervised method for anomaly localization in volumetric brain MRIs. Building upon the MSMA framework, our approach incorporates spatial information and conditional likelihoods to enhance anomaly detection capabilities. We employ a flexible normalizing flow model conditioned on patch positions and global image features to estimate patch-wise anomaly scores. The method is evaluated on a dataset of 1,650 T1- and T2-weighted brain MRIs from typically developing children, with simulated lesions added to the test set. Spatial-MSMA significantly outperforms existing methods, including reconstruction-based, generative-based, and interpretation-based approaches, in lesion detection and segmentation tasks. Our model achieves superior performance in both distance-based metrics (99th percentile Hausdorff Distance: $7.05 \pm 0.61$, Mean Surface Distance: $2.10 \pm 0.43$) and component-wise metrics (True Positive Rate: $0.83 \pm 0.01$, Positive Predictive Value: $0.96 \pm 0.01$). These results demonstrate Spatial-MSMA's potential for accurate and interpretable anomaly localization in medical imaging, with implications for improved diagnosis and treatment planning in clinical settings.
 
	Our code is available at~\url{https://github.com/ahsanMah/sade/}.
\end{abstract}


\section{Introduction}

Multiscale score matching analysis  (MSMA) was introduced by~\cite{mahmood2021multiscale} as a novel methodology for sample-wise anomaly detection. The method learns the distribution of the magnitudes of the gradient of log-likelihoods i.e. the norm of the score function: $\norm{s(x)} = \norm{ \nabla_x \log p(x)}$. 
Sample-wise anomaly detection operates on the full input and makes binary predictions on whether a sample is anomalous. It does not 
 provide information about the input components that led to that assessment.

However, it is often desirable to identify the specific regions within an image that are contributing to its atypicality. Such localization allows for increased model interpretability as well as directing future investigation. For instance, in healthcare, the ability to interpret a model's prediction empowers medical practitioners to visually corroborate the identified regions of interest. Interpretation may also pave the way for novel insights into the disease. Furthermore, localizing anomalies enables targeted diagnosis and intervention planning based on the factors contributing to the detected outlier.

 To address this, we propose Spatial-MSMA, an extension of MSMA that incorporates spatial information to enable precise anomaly localization. Spatial-MSMA leverages the concept of patch-wise analysis, considering not only the content of individual patches but also their spatial context within the image. By modeling conditional likelihoods that account for patch position and global image features, Spatial-MSMA provides a more nuanced and interpretable approach to anomaly detection and localization. For example, Spatial-MSMA is capable of highlighting brain lesions even though it was never trained with labeled data.

\section{Background}
\label{ch:background}

Denoising score matching and MSMA (Multiscale Score Matching Analysis), play a pivotal role in Spatial-MSMA. This section elucidates score matching, its extensions such as noise-conditioned score matching, and its application in MSMA.

\subsection{Score Matching}
A score is defined as the gradient of the log probability density, with respect to the data. Conceptually, a score is a vector field that points in the direction where the log density grows the most. 
\cite{hyvarinen2005} introduced score matching as a means of computing the parameters of an unnormalized probability density model. The authors proved the remarkable property that learning the score involves the gradient of the score function itself as shown in Equation~\ref{eq:implicit_sm}. Following the naming scheme used in \cite{vincent2011connection}, this objective is called Implicit Score Matching.

\begin{align}
\label{eq:implicit_sm}
    J_{ISM}(\theta) &= \mathbb{E}_{x \sim p(x)} \frac{1}{2} \s{ \norm{s_\theta(x) - \nabla_x \log p(x) }^2 } \\
    &= \mathbb{E}_{x \sim p(x)} \s{ \norm{s_\theta(x)}^2 + \sum^{d}_{i=1}{\partial{x_{i}} s(x_{i}) } }
\end{align}

\subsection*{Denoising Score Matching}

\cite{vincent2011connection} formalized a connection between denoising autoencoders and score matching, and proposed the denoising score matching (DSM) objective. The authors noted how \cite{hyvarinen2005} had suggested the possibility of an alternate score matching objective; one that was based on regressing against the data gradients of a Parzen window density estimator. This so-called Explicit Score Matching objective is shown in \eqnref{eq:explicit_sm}.

\begin{align}
\label{eq:explicit_sm}
    J_{ESM}(\theta) &= \mathbb{E}_{x \sim q_{\sigma}(x)} \frac{1}{2} \s{ \norm{s_\theta(x) - \nabla_x \log q_{\sigma}(x) }^2 }
\end{align}

\cite{vincent2011connection} proved that under certain regularity conditions,\footnote{For any window size $\sigma > 0$, the kernel $q_{\sigma}$ is differentiable, converges to 0 at infinity, and has a finite gradient norm}
the Parzen window based objective is equivalent to the original objective proposed by \cite{hyvarinen2005} in \eqnref{eq:implicit_sm}

Taking it one step further, assume the Parzen density estimate is chosen to estimate the joint density of clean and \textit{corrupted} samples $(x, \tilde{x})$ i.e. $q_{\sigma}(x, \tilde{x}) = q_{\sigma}(x | \tilde{x} ) p(x) $. Thus, the DSM objective is simply:
\begin{align}
\label{eq:denoising_sm}
    J_{DSM}(\theta) &= \mathbb{E}_{\tilde{x}\sim q_{\sigma}(x, \tilde{x})} \frac{1}{2} \s{ \norm{s_\theta(\tilde{x}) - \nabla_{\tilde{x}} \log q_{\sigma}(\tilde{x} | x ) }^2 }
\end{align}

DSM mitigates the need for computing second order gradients as is the case for \eqnref{eq:implicit_sm}. Furthermore, if $q_{\sigma}$ is set as the Gaussian kernel $\mathcal{N}(\tilde{x} |  x,\,\sigma^{2}I)$, then $\nabla_x \log q_{\sigma}(\tilde{x}) = \frac{(x - \tilde{x})}{\sigma}$. One can now see the connection between score matching and the denoising autoencoder objective (when using a Gaussian kernel). The score model is effectively being trained to estimate the \textit{noise} that was added to the image.

It should be emphasized that while~\cite{vincent2011connection} and many subsequent works \cite{Song2019,song2020improved,song2020score}, use the Gaussian distribution in DSM, the proof for the validity of the objective by~\cite{vincent2011connection} holds for \textit{any} differentiable noise distribution.

\subsection*{Noise Conditioned Score Matching}
\label{NCSM}

\cite{Song2019} extended the DSM objective in \eqnref{eq:denoising_sm} to incorporate multiple scales $\sigma$, and train a so-called Noise Conditioned Score Network (NCSN). The authors further outlined an iterative sampling algorithm, dubbed annealed Langevin dynamics, enabling the score network to be employed as a deep generative model. Let $\{\sigma_i\}_{i=1}^L$ be a positive geometric sequence that satisfies $\frac{\sigma_1}{\sigma_2} = ... = \frac{\sigma_{L-1}}{\sigma_{L}} >  1$. NCSN is a conditional network, $s_{\theta}(x,\sigma)$, trained to jointly estimate scores for various levels of noise $\sigma_i$ such that $\forall \sigma \in \{\sigma_i\}_{i=1}^L: s_{\theta}(x,\sigma) \approx \nabla_x \log q_{\sigma}(x)$. In \cite{Song2019}, the conditioning information is explicitly provided via a one-hot vector denoting the noise level used to perturb the data. The network is then trained via a modified denoising score matching objective as shown in \eqnref{eq:ncsn_dsm}.
\begin{equation}
\label{eq:ncsn_dsm}
J_{NCSN}(\theta) = \frac{1}{L} \sum_{i=1}^L \lambda(\sigma_i)
\s{\frac{1}{2} \displaystyle \text{~} \mathbb{E}_{\tilde{x}\sim q_{\sigma_i} (\tilde{x}|x) p_{\text{data}}(x)} \s{ \norm{ s_{\theta}(\tilde{x}, \sigma_i) + (\frac{\tilde{x} - x}{\sigma_i^2}) }^2_2  } }
\end{equation}


\subsection*{Connecting Score Matching to Diffusion Models}

In a follow-up work,~\cite{song2020score} described a connection between noise-conditioned score matching and diffusion models~\cite{sohl2015deep}. The connection is presented under the lens of generative modeling, and provides a framework which unifies Markov-based and continuous-time diffusion models. The key insight is that successive perturbation of a data point using a scale-dependent noise distribution (as done in NCSNs), follows a Stochastic Differential Equation (SDE). Thus, the `forward' diffusion process can be modeled as an SDE. If one has access to the scores at each time point, it is possible to construct a reverse-time SDE as proved by~\cite{anderson1982reverse}. These reverse-SDEs can be numerically solved using any differential equation solver, only requiring access to the score function. The authors were able to use this formulation to generate images that surpassed the state-of-the-art generative models. This work was seminal in the development of future diffusion models.

\subsection*{Continuous-Time Score Matching}
More relevant to this research,~\cite{song2020score} enabled a continuous relaxation to the discretized nature of noise conditioned score matching. Defining the noising process as a continuous forward SDE alleviates the need to predetermine the number of noise scales (as was the case for NCSN models). For generative models, this translates to faster sampling as one can control the number of gradient steps to take. For this work, it gives us the ability to observe different noise scales at test time while maintaining the advantages of using multiple noise scales during training. Namely, it forces the model to learn smoother transitions between noise scales, which helps test time generalizability.

\subsection{Multiscale Score Matching Analysis}
\label{sec:msma}

While~\cite{Song2019} demonstrated the generative capabilities of NCSNs, ~\cite{mahmood2021multiscale} outlined how 
these networks can be repurposed for outlier detection. Their methodology, Multiscale Score Matching Analysis (MSMA), incorporates noisy score estimators to separate in- and out-of-distribution (OOD) points. Recall that a score is the gradient of the likelihood. A typical point, residing in a space of high probability density will need to take a small gradient step in order to improve its likelihood. Conversely, a point further away from the typical region (an outlier) will need to take a comparatively larger gradient step towards the high density region.  When we have multiple noisy score estimates, it is difficult to know apriori which noise scale accurately represents the gradient of the outliers. However, ~\cite{mahmood2021multiscale}, showed that learning the typical space of score-norms for all noise levels is sufficient to identify anomalies.

Concretely, assume we have a score estimator that is trained on $L$ noise levels and a set of inlier samples $X_{\text{IN}}$. Computing the inlier score estimates for all noise levels and taking the L2-norms across the input dimensions results in an $L$-dimensional feature vector for each input sample: $[\norm{  s (X_{\text{IN}}, \sigma_1) }_2^2, ... , \norm{  s (X_{\text{IN}}, \sigma_L) }_2^2 ]$. The authors in~\cite{mahmood2021multiscale} argue that inliers tend to concentrate in this multiscale score-norm embedding space. It follows that one could train an auxiliary model (such as a clustering model or a density estimator) to learn this score-norm space of inliers. At test time, the output of the auxiliary model (e.g. likelihoods in the case of density estimators) is used as an anomaly score. Results in~\cite{mahmood2021multiscale} show MSMA to be effective at identifying OOD samples in image datasets (e.g. CIFAR-10 as inliers and SVHN as OOD).

Our work builds on the MSMA methodology for anomaly detection. We will localize anomalies within the image by estimating the likelihoods of \textit{patch} score-norms. This requires careful consideration of model architecture and the likelihood method applied, as we will discuss in Section~\ref{sec:spatial-msma}.

\subsection{Normalizing Flows}
\label{sec:normflows_intro}
Spatial-MSMA employs deep likelihood (generative) models to estimate the likelihoods of the patch score norms. Normalizing flow models are a flexible class of generative models which can efficiently estimate the likelihood of a sample by usign the change of variable formula. They utilize invertible transformations to project the data into the space of a predefined base distribution, such as a Standard Gaussian. Examples of deep normalizing flows are models such as RealNVP~\citep{dinh2017density} or Neural Spline Flows~\citep{NEURIPS2019_7ac71d43}, which are typically trained via the maximum likelihood objective. However, these models fail to detect outlying samples much like deep autoregressive models. The work by ~\cite{normflow_fails} investigated the failure cases of flow models. The authors analyses suggest that flow models encode the visual appearance directly, without learning any semantic content. The anomaly detection performance of flow-based models can improve if they are trained on high-level semantic representations (e.g. from a pretrained neural network) rather than the raw images themselves. 

\section{Related Works}
\label{sec:related}

There are many existing approaches for anomaly detection. We will focus on a few that are commonly employed for medical imaging. These approaches can be organized into four broad categories as described below. 

\subsection{Reconstruction based Approaches}
Reconstruction-based anomaly detectors are trained to produce typical counterparts (so-called reconstructions) of anomalous images. The methods may take some form of a deep autoencoder~\citep{dae-kascenas22a,baur2021}, trained with a reconstruction error objective such as mean squared error. At test time, the models are presumed to output an anomaly-free image, with the \textit{reconstruction error} as the metric of atypicality. A known drawback of these models is the lack of specificity in their detection. As no reconstruction is pixel-perfect (especially in terms of image intensities), the output error maps have significant false-positives~\citep{baur2021}. Another drawback of autoencoders is that as their reconstruction abilities improve, their anomaly detection capabilities decrease as the models are better at reconstructing the anomalies.

\subsection{Generative Modeling based Approaches}
While reconstruction-based methods directly learn to reproduce typical samples, generative modeling approaches learn the underlying data distribution. These models can then use iterative sampling techniques to gradually modify the input image into a in-distribution sample. There are two main methods proposed to employ generative models for removing anomalies: \textit{imputation-based} such as ~\cite{liuUnsupervisedOutofDistributionDetection2023} or \textit{restoration-based} such as ~\cite{pmlr-v102-you19a}.

Unlike reconstruction-based approaches that attempt to recreate the entire input, imputation-based methods utilize masking strategies to selectively reconstruct certain regions of the image. Multiple passes are used to mask out different regions of the image. By focusing the network on only the unmasked regions, this approach can potentially lead to more precise anomaly localization. Restoration-based approaches, on the other hand, modify the entire image using the original image as the starting point in the sampling procedure, which can be more effective for global anomalies. The key advantage of generative modeling approaches over reconstruction-based methods is their ability to capture the probabilistic nature of the data distribution. This allows them to potentially handle a wider range of anomalies and provide more nuanced anomaly scores. However, they can be more computationally intensive and may require more complex training procedures compared to simpler reconstruction-based autoencoders.

Recently, owing to the success of score-based diffusion models, much of the research has focused on using diffusion models as the generative model (replacing GANs of yesteryear)~\citep{wyattAnoddpmAnomalyDetection2022,pinaya2022fast,liuUnsupervisedOutofDistributionDetection2023,behrendtPatchedDiffusionModels2023}. These models provide slight modifications to the diffusion sampling process, starting the generation from an input image rather than random noise. The sampling process in these generative models typically involves adding noise to the input image and iteratively denoising the sample to generate an anomaly-free counterpart of the original image. Once the cleaned sample is generated, a voxel-wise difference between the input and its anomaly-free counterpart is used as the anomaly score. This process differs from reconstruction-based methods, which directly output the reconstructed image in a single forward pass, without adding any noise to the input. The main differentiating factors between these generative modeling methods are the hyperparameters used for training and the sampling strategies used during inference, which can significantly impact their performance and computational efficiency.

\subsection{Feature Embedding based Approaches}
Some methods aim to detect anomalies in a learned embedding space. The feature embeddings are computed by a neural network trained on the typical samples. At test time, it is assumed that the model will output feature embeddings that are \textit{close} to the feature embeddings of the training population if the sample is an inlier and \textit{away} otherwise. A popular method in this category is the Student-Teacher architecture anomaly detector by~\cite{bergmann2020uninformed}. In this setting, we have two models: a high parameter-count Teacher network and low parameter-count Student network. The student model is considered to be a ``weaker" version of the Teacher, and is trained using neural network distillation techniques. It is assumed that the Student model will fail to generalize to unseen datasets, producing a discrepancy between the teacher's features and that of the student. This discrepancy is used to produce an anomaly score.

\subsection{Attribution based Approaches}
Certain anomaly detection techniques draw on insights from interpretability research. The task is to identify features of the data that contribute to the model's output. The identified features are often assigned a score relative to their importance, as determined by the rules of the interpretation technique. Examples of such methods include SHAP~\citep{NIPS2017_7062}, Saliency Maps~\citep{Simonyan2013DeepIC}, and GradCAM~\citep{gradcam}.

SHAP (SHapley Additive exPlanations) is a game-theoretic approach to explain the output of any machine learning model. SHAP values attribute the prediction of an instance to the different features, highlighting the positive or negative impact of each feature. Saliency Maps are a visualization technique that highlight areas of an input image that most influence the output of a network. The saliency map is computed by taking the gradient of the output with respect to the input image. Areas with high gradient values correspond to regions in the input that have a significant impact on the model's prediction. GradCAM (Gradient-weighted Class Activation Mapping) is an extension of saliency maps that computes gradients with respect to feature vectors of an image rather than the image itself.

All of the mentioned attribution-based approaches aim to identify the features or input regions that contribute most to the model's predictions. This information can be used for anomaly detection, as it can localize the patches that lead a model to classify an instance as an anomaly.

\subsection{Situating Spatial-MSMA within Existing Works}
Spatial-MSMA addresses key limitations of existing anomaly localization techniques. Unlike reconstruction-based approaches, it avoids the pitfall of decreasing detection capabilities as reconstruction quality improves. In contrast to generative modeling methods, Spatial-MSMA doesn't require complex sampling procedures or modifications to pretrained models. It surpasses feature embedding techniques by considering both local and global context, potentially leading to more nuanced anomaly detection. Unlike attribution-based methods that often require labeled data or focus solely on model interpretability, Spatial-MSMA offers unsupervised learning with built-in localization capabilities. By leveraging conditional likelihoods and spatial information, Spatial-MSMA is expected to provide more accurate and interpretable anomaly localization, making it particularly valuable for applications in healthcare and manufacturing where precise identification of atypical regions is crucial.

\section{Spatial-MSMA: Incorporating Spatial Information into MSMA}
\label{sec:spatial-msma}

\begin{figure}[h]
    \centering
    \includegraphics[width=\textwidth]{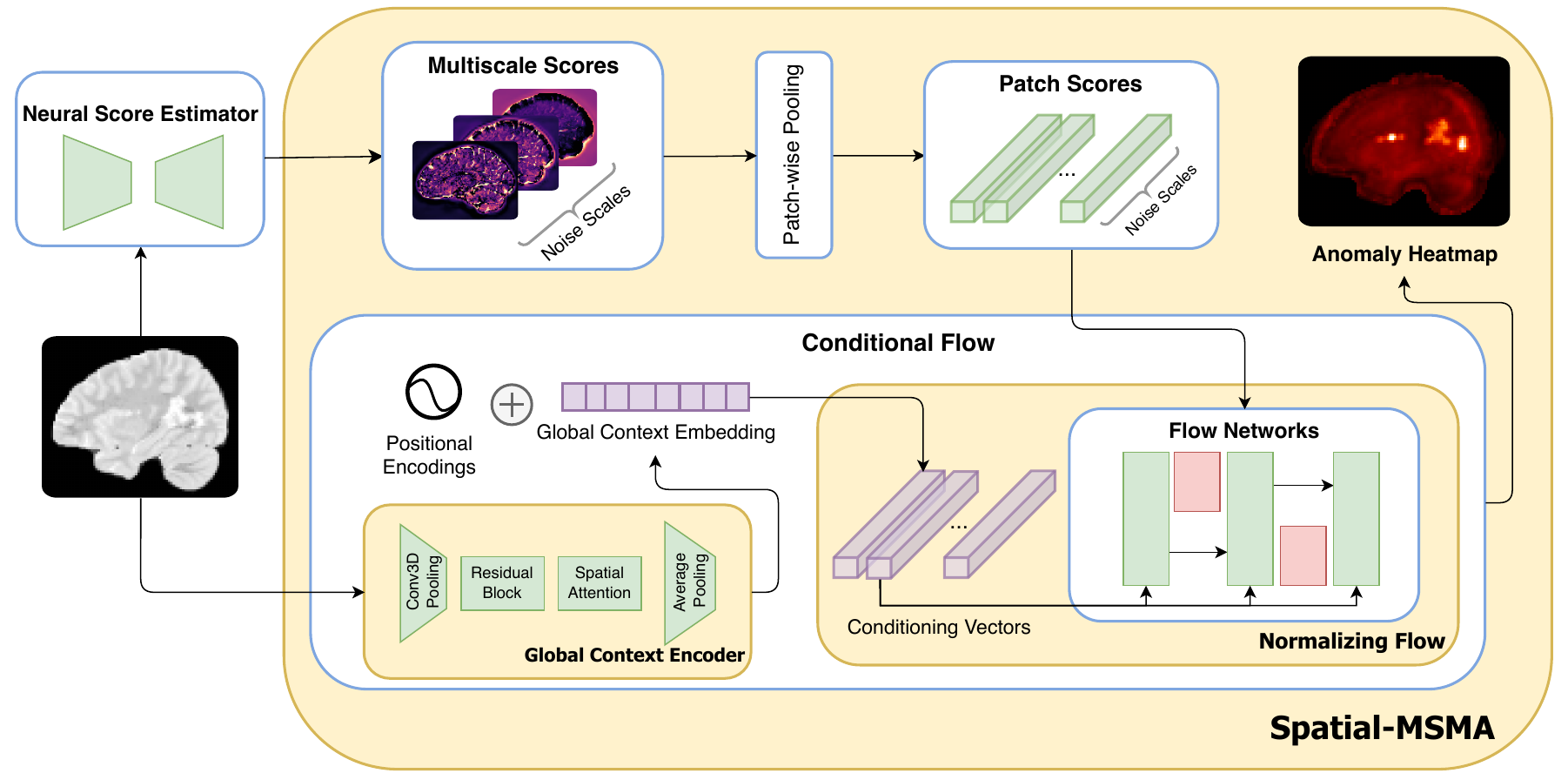}
    \caption{Overview of Spatial-MSMA. A neural score estimator produces score tensors at multiple noise scales. The score tensors are divided into patches and processed by a conditional flow to estimate patch-wise anomaly scores. Global image features are extracted by a convolutional network and combined with positional encodings corresponding to each patch location, resulting in a conditioning vector per patch. The patch score norms and conditioning vectors are fed into a normalizing flow model with conditional coupling blocks. The result is a negative likelihood heatmap that highlights anomalous patches within the image. Spatial-MSMA thus enables precise localization of anomalies based on the patch scores and their spatial context.}
    \label{fig:overview}
\end{figure}

The basic assumption of MSMA, as explained in Section~\ref{sec:msma}, is that inliers will occupy distinct regions in the score-norm space. At test time, we ask the question: Does the given sample belong to the inliers? MSMA consequently estimates the likelihood of a sample belonging to the inlier region in the score-norm space. Thus, MSMA looks at the data samples holistically, i.e. it considers the entire set of features available (e.g. all the pixels in an image). 

However, we note that MSMA is also amenable for \textit{subsets} of features. For instance, we may divide an image into patches and consider the score-norms of each patch independently. Now, we can ask the question: Does this \textit{patch} belong to the inliers? As before, MSMA will output a likelihood estimate of a test patch belonging to the inliers, but this time \textit{only} considering information present at the given patch location.

It is possible to naively extend MSMA to consider patches. One can decompose the image into a regular grid, and train an independent MSMA model for each grid location. One may even reduce computational costs by running training/inference in parallel for each patch location. However, while this approach is straightforward, it leaves much room for improvement.

Namely, we can leverage \textit{spatial locality}: the notion that neighbouring image patches are highly correlated. Furthermore, even patches which are spatially apart may depend on each other. Consider an image of a face.
Observing patches of the left eye gives us rich information about what we may observe in the location of the \textit{right} eye, even if the location of the right eye is distal to the left.
One can incorporate this information into the decision making process to reason about the typicality of a queried patch. For instance, observing a brown-colored right eye is typical. However, observing a brown colored right eye \textit{given} a black-colored left eye, is atypical.

\subsection{Modeling Conditional Likelihoods}
Following the motivation above, one can employ a conditional model where in addition to the contents of a patch, its position and surrounding context are also taken into account. As such, we posit to use a conditional likelihood model as the basis of our patch-based anomaly detector.

Concretely, the model will be conditioned on the patch position and the image features. Let $ s_p = \{s(x_p)\}_{i=1}^{L}$ be the multi-scale score tensor for a given patch $x_p$ at location $p$, belonging to the image $x$. Let $h(x)$ are the feature vectors of the image $x$ computed by a convolutional network $h$. We propose to estimate the conditional likelihood model $p(s_p | p, h(x))$. Both the normalizing flows and the context-encoding convolutional network are trained via the Maximum Likelihood Estimation (MLE) objective. As this model will output likelihoods of score-norms for each patch conditioned on the surrounding spatial information, the model is called Spatial-MSMA.

Spatial-MSMA uses a flexible class of likelihood estimators called normalizing flows introduced in Section~\ref{sec:normflows_intro}. The patch locations are modeled via sinusoidal positional embeddings, commonly used in Transformer models~\citep{vaswani}. In order to capture global image context, the original image is passed through a convolutional network with a large receptive field. The resulting feature embeddings are concatenated with the positional embeddings and fed into the flow model as contextual information. Recall that the feature embeddings are learned, as the convolutional network is trainable and will be updated in every backwards pass. The positional embeddings are fixed and not trained. The flow model then transforms the input samples, following the constraints of the change-of-variable technique. Finally the flow estimates likelihoods of the transformed samples via a trainable conditional Gaussian Mixture Model (GMM). 

\section{Lesion Detection in Volumetric Brain MRIs}

To illustrate the anomaly detection capabilities of Spatial-MSMA,  will consider the task of detecting lesions in medical images.We choose this to reflect the real world usecase of automatic detection and segmentation of pathologies. Recall, the model will be trained without any labels and will not be fed any prior knowledge about the type of anomalies at test time.  

To minimize confounding factors that can be introduced due to a distributional shift between the training (healthy) and testing (lesioned) populations, the anomalies will be simulated on a held out inlier test set, ensuring that the introduced anomalies are the \textit{primary} factor differentiating the test set from the inlier population.

\subsection{Constructing a Healthy Population}

Our inlying, healthy population will comprise of typically developing school-age children. We chose this cohort due to the availability of public datasets within this demographic. Speicifically, we retrieved data from two studies: the Adolescent Brain Cognitive Development (ABCD) Study~\citep{Casey2018adolescent}, and the Human Connectome Project Development (HCP-D) Study~\citep{somerville2018lifespan}. Samples from these studies were preprocessed to remove any outliers. To keep the inlier cohort as nominal as possible, we used the Child Behavior Checklist (CBCL)~\cite{achenbachChildBehaviorChecklist1999} scores as our filtering mechanism. This checklist assesses the behavior and emotional competencies of children. Children with behavioral problems tend to score high on this test. For our analysis, all children that scored above a t-score of 66 ($\sim 95$-th percentile) in the summary scores as well as \textit{any} of the subscores were removed. Note that this is more conservative than only using the summary scales. The data was then split into an 80/10/10 train/validation/test split. Our processing resulted in 1320 training, 165 validation, and 165 testing samples. 

We use both T1-weighted and T2-weighted images. As the images are high-resolution 3D MRIs, they require a lot of GPU memory during training. In order to fit a batch size of 4 per GPU, the images were downsampled to a pixel spacing of 2mm isotropic. They were further cropped by the largest brain mask, computed from the training data. After some padding to make the images multiples of 2, the resulting 3D volume was of size 96x112x80. 

\subsection{Simulating Lesions}
The lesions were simulated using a lesion simulator tool~\cite{Filho_2019}, available as the MSLesionSimulator extension~\footnote{https://www.slicer.org/wiki/Documentation/Nightly/Modules/MSLesionSimulator} of the Slicer3D software package~\cite{fedorov3DSlicerImage2012}. The lesion load parameter was set to 20 and the rest of the hyperparameters were kept at their default values. A post processing step was performed to enhance the lesion intensity by a factor of $1.5$. The lesions were generated on the test set.

\subsection*{Training Details}

The score-norms were retrieved from a score-based diffusion model, using a 3D convolutional UNet-like architecture. We followed the SDE formulations by~\cite{song2020score}, using the `Variance-Exploding' SDE with 2000 timesteps. The minimum sigma was set to 0.06, which is the average standard deviation of the image intensities. This is done so that, at minimum, the model is able to capture the intensity variation of within an image. Following the suggestion of ~\cite{song2020score}, the maximum sigma was set to 545.0 which is the 99-th percentile of the pairwise distance in the training set. This is done to allow the largest noise distribution to maximally cover the support of the data distribution i.e. $p_{\sigma_\text{max}}(x) \approx \mathcal{N}(x| 0, \sigma_\text{max}^2I)$. The model was trained for 1.5 million iterations, by which point the validation loss had started to flatten out. The batch size was doubled at roughly the half way point during training. This is a simple yet effective method proposed by~\cite{le2018dont}, to effectively anneal the learning rate without having to use a decay schedule. The authors also reported that increasing the batch size reduced the number of parameter updates required to reach the same test accuracies when compared to strategies for decaying the learning rate.

During inference, the voxel-wise anomaly scores are first brain masked followed by thresholding. The threshold is determined for each sample by searching for the threshold that gives the lowest symmetric mean surface distance between the ground truth and the post-threshold segmentation. Searching for a threshold like this is common practice in evaluating anomaly detectors~\cite{baur_deep_2019}. The segmentations are post-processed by removing connected components of size less than 3 voxels (using a connectivity of 1). The remaining segmentation mask is dilated via a disk of radius 1 as the structuring element. Note that this inference procedure is performed for all methods tested in the experiment.

\subsection{Baseline Methodologies}

Spatial-MSMA was compared to a selection of models that encompass a broad range of anomaly detection methodologies that have been successfully used in the medical imaging field. Namely, the baselines represent reconstruction-based, generative-based, and interpretation-based methods.

For the reconstruction-based baseline, we chose an autoencoder model by~\cite{aelu2023} owing to its success on volumetric brain MRIs. The model uses a ResNet-like architecture is trained using a reconstruction objective based on a Mean Squared Error (MSE). The authors also provide a publicly available implementation. This method is denoted as AE in Table~\ref{lesion_results}.

Two generative-model based approaches were also included in the comparison. First is an imputation-based approach inspired by~\cite{liuUnsupervisedOutofDistributionDetection2023} which uses a checkerboard mask to in-paint different regions of the image (denoted as Inpaint in Table~\ref{lesion_results}). This method performs multiple runs of imputation, alternating the checkerboard pattern each time and computing the average error across all runs. Second is a restoration-based approach (denoted as Restoration in Table~\ref{lesion_results}), which first adds noise to the image and then invokes the sampling procedure of the diffusion model to iteratively generate the restored counterpart. Following~\cite{wyattAnoddpmAnomalyDetection2022}, the sampling procedure was initiated from 1/4th of the original timesteps. However, unlike~\cite{wyattAnoddpmAnomalyDetection2022}, we did not use Simplex noise during training/inference as work by ~\cite{kascenas2023} has shown that it may not be necessary (and sometimes detrimental). Note that these technique are agnostic to the diffusion framework, which allows us to use the same diffusion model that was used as the backbone of Spatial-MSMA. This keeps the comparison fair and limits confounding factors as the diffusion model was trained to convergence and is able to generate realisitic looking samples.

Lastly, GradCAM was included as a representative of attribution-based approaches. Specifically, we used Guided-GradCAM~\citep{Selvaraju2016GradCAMVE} which combines saliency maps (with some modifications) and GradCAM to give superior results to vanilla GradCAM. The gradients were computed using the outputs of a non-spatial MSMA. Concretely, a GMM was trained on the whole-image score norms and the GradCAM gradients were computed using the negative likelihood estimates. This corresponds to computing voxel-wise attribution maps for an MSMA anomaly score. Thus, the method is denoted as GradCAM-MSMA in Table~\ref{lesion_results}.

\subsection{Segmentation Metrics for Analysis}
We chose mean surface distance (MSD) and the Hausdorff distance (HD) as the primary segmentation metrics for comparison. MSD calculates the average distance between the surfaces of the predictions and ground truth, providing a measure of overall segmentation accuracy. The Hausdorff distance represents the maximum distance between the two surfaces, capturing the worst-case scenario. We use the 99th percentile of the Hausdorff distance to mitigate the impact of outliers. These metrics are less biased towards over-segmentations compared to the more popular Dice score, making them particularly suitable for anomaly detection tasks where false positives can be problematic. Both distances are computed in a directed manner, i.e., from the ground truth to the prediction.

In addition to distance metrics, we computed component-wise metrics to assess the model's ability to identify distinct anomalous regions. Connected components were determined from the voxel-wise segmentation masks using an 8-connectivity-neighborhood (including diagonals). We assign a \textit{true positive} (TP) label to a predicted component that overlaps with any ground-truth component at any voxel location. The absence of any overlap is tallied as a \textit{false positive} (FP). \textit{False negatives} (FN) are ground-truth components with no corresponding prediction overlap. Table~\ref{lesion_results} reports the True Positive Rate (TPR = TP/(TP+FN), measuring sensitivity) and the Positive Predictive Value (PPV = TP/(TP+FP), measuring precision). These component-wise metrics provide insights into the model's performance in detecting and delineating individual anomalous regions.

\begin{figure}[htb!]
\centering
\includegraphics[width=\textwidth]{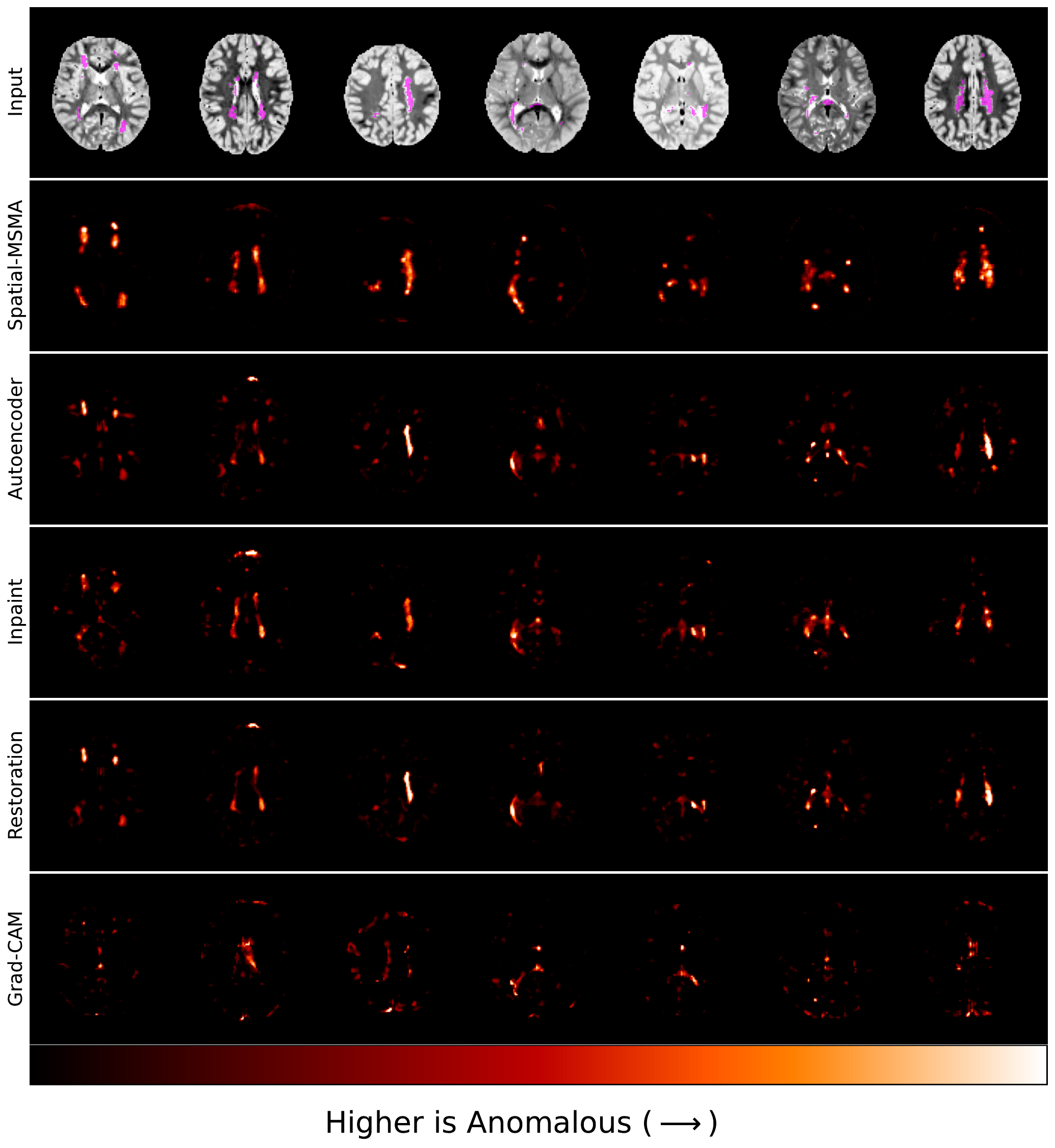}
\caption{Qualitative comparison of anomaly heatmaps across different methods. The first row shows random axial slices of the volumetric input samples. The lesions are highlighted in magenta. Each column is a slice from random individuals. Note how Spatial-MSMA consistently detects all the lesions in the image, while other methods tend to miss smaller lesions.}
\label{fig:lesion-comp}
\end{figure}

\subsection{Results}

Table~\ref{lesion_results} presents the segmentation performance of Spatial-MSMA compared to baseline methodologies across various metrics. We report the mean across (165) test samples alongside the standard error for each metric. Note that due to the relatively small size of the lesions, the segmentation task was difficult for all models. However, the results demonstrate Spatial-MSMA's superior performance in lesion detection and localization tasks. 

Spatial-MSMA achieved the lowest 99th percentile Hausdorff Distance (99-HD) of 7.05 and Mean Surface Distance (MSD) of 2.10. These metrics reflect the model's ability to accurately delineate lesion boundaries. The significantly lower distances compared to baselines (e.g., Restoration: 99-HD of 8.67, MSD of 2.68) indicate that Spatial-MSMA produces tighter and more precise segmentations around anomalies.

The component-wise metrics reflect the sensitivity of the model to anomalous regions in the image, regardless of size. 
Spatial-MSMA exhibited exceptional performance in detecting individual lesion components, with a True Positive Rate (TPR) of 0.83 and a Positive Predictive Value (PPV) of 0.96. Recall that the maximum possible for either metric is 1.0. This high TPR indicates that Spatial-MSMA successfully identified 83\% of all lesions, significantly outperforming the next best method (Restoration, TPR: 0.68). The high PPV of 0.96 suggests that Spatial-MSMA has a very low false positive rate, a crucial factor in clinical applications where false alarms can lead to unnecessary interventions or patient anxiety.

Figure~\ref{fig:lesion-comp} provides a visual comparison of anomaly heatmaps generated by different methods. The plotted heatmaps are clipped at the 90th percentile for each sample i.e. the range represents the top-$10\%$ of the anomaly scores.
Spatial-MSMA consistently detects all lesions in the image, including smaller ones that other methods often miss. This is particularly evident in the third column, where Spatial-MSMA correctly identifies a small lesion (bottom left) that goes undetected by other approaches. The heatmaps also demonstrate Spatial-MSMA's ability to provide more focused and accurate localization, with less diffuse activation around the lesions compared to methods like GradCAM. Note that while Spatial-MSMA tends towards over-segmentation, it manages to detect most if not all the lesions. Other baselines such as Inpaint, and Restoration are overly biased towards larger anomalies, and often fail to detect smaller lesions.

\begin{table*}[!ht]
\centering
\caption{Segmentation metrics for lesion detection. Each model was trained only on inlier samples only, and tested on the same 165 test samples. We report the mean across the test samples alongside the standard errors. Right column shows distance based metrics: 99th-percentile of the Hausdorff Distance (99-HD) and Mean Surface Distance (MSD). Right column shows component-wise metrics: True Positive Rate (TPR = TP/(TP+FN)) and Positive Predictive Value (PPV = TP/(TP+FP)). Spatial-MSMA significantly outperforms the baseline methodologies, especially for component-wise metrics. }
\begin{tabular}{lrr|rr}
{} &        99-HD~$\downarrow$~~&      MSD~$\downarrow$~~~~~&        TPR~$\uparrow$~~~~&             PPV~$\uparrow$~~~~~\\
\midrule
AE~\citep{aelu2023} &  12.27$\pm$~0.51 &  3.63$\pm$~0.35 &  0.44$\pm$~0.02 &  0.19$\pm$~0.01 \\
Inpaint~\citep{liuUnsupervisedOutofDistributionDetection2023}      &  13.26$\pm$~0.50 &  3.71$\pm$~0.27 &  0.63$\pm$~0.02 &  0.50$\pm$~0.02 \\
Restoration~\citep{wyattAnoddpmAnomalyDetection2022}  &   8.67$\pm$~0.53 &  2.68$\pm$~0.36 &  0.68$\pm$~0.02 &  0.17$\pm$~0.01 \\
GradCAM-MSMA &  12.68$\pm$~0.54 &  3.75$\pm$~0.37 &  0.43$\pm$~0.02 &  0.16$\pm$~0.01 \\
Spatial-MSMA &  \textbf{7.05~$\pm$~0.61} &  \textbf{2.10~$\pm$~0.43} &  \textbf{0.83~$\pm$~0.01} &  \textbf{0.96~$\pm$~0.01} \\
\end{tabular}

\label{lesion_results}
\end{table*}

\section{Discussion and Limitations}
The results of our study demonstrate the effectiveness of Spatial-MSMA in addressing the challenging task of unsupervised anomaly detection and localization in medical imaging. The superior performance of Spatial-MSMA across various metrics highlights its potential as a powerful tool for automated lesion detection in brain MRIs. The difficulty of unsupervised anomaly detection in medical imaging cannot be overstated. Unlike supervised learning approaches that rely on large datasets of labeled abnormalities, unsupervised methods must learn to identify anomalies without prior knowledge of what constitutes an anomaly. This is particularly challenging in medical contexts where anomalies can be subtle, diverse, and often mimicked by normal anatomical variations. Spatial-MSMA's ability to achieve high detection rates and low false positive rates in this unsupervised setting is therefore especially noteworthy.

Furthermore, the incorporation of spatial information and conditional likelihoods in Spatial-MSMA addresses key limitations of previous approaches. By considering both local patch content and global image context, Spatial-MSMA can better distinguish between true anomalies and normal variations that may appear anomalous when viewed in isolation. This is evident in the model's ability to detect small lesions that other methods miss, as well as its high precision in avoiding false positives.

Note that this high precision (as illustrated by the TPR and PPV achieved by Spatial-MSMA) has significant implications for clinical applications. A high true positive rate ensures that clinicians are alerted to potential abnormalities, while a high positive predictive value minimizes false alarms. This balance is crucial in medical settings where both missed diagnoses and overdiagnosis can have serious consequences for patient care and resource allocation.

However, despite its strong performance, Spatial-MSMA is not without limitations. The model's tendency towards slight over-segmentation, while preferable to under-segmentation in many clinical contexts, could be further refined. Future work could explore ways to fine-tune the model's sensitivity to strike an even better balance between detection and precision. Another area of improvement is the computational resources required for both training and inference when compared to more efficient baselines such as an autoencoder. This is due to the high parameter count of the underlying score-based diffusion models. Note that Spatial-MSMA is still an order of magnitude faster than the iterative sampling based solutions such as AnoDDPM (Inpaint) ~\cite{wyattAnoddpmAnomalyDetection2022}.


\section{Conclusion}
This work introduced Spatial-MSMA, a novel extension of Multiscale Score Matching Analysis that incorporates spatial information for improved anomaly detection and localization. By leveraging conditional likelihoods and patch-based analysis, Spatial-MSMA demonstrated superior performance in detecting and localizing simulated lesions in volumetric brain MRIs compared to several state-of-the-art baselines.
Our results showed that Spatial-MSMA significantly outperformed existing methods across multiple metrics, including mean surface distance, Hausdorff distance, true positive rate, and positive predictive value. The model's ability to detect lesions of varying sizes while maintaining a low false positive rate highlights its potential as a powerful tool for unsupervised anomaly detection in medical imaging.

Future work could explore several avenues such as extending the model to handle multi-modal imaging data, and unifying the learning objectives so as to train the score-estimator and the normalizing flow model in an end-to-end fashion. One may also explore the application of Spatial-MSMA to other domains beyond medical imaging, such as manufacturing or satellite imagery, where spatial context is crucial for anomaly detection.

\acks{This work was supported by grants R01-HD055741, U54HDO79124, HD053000, MH070890, MH111944, EB021391, P50HD103573, MH115046, 1R01AA02687901A1, 1OT2OD032581-02-321, NSF IIS2133595, and NSF DMS2324394.

HCPD data generation was supported by grants U01MH109589 and U01MH109589-S1.
}

%
\ethics{The work follows appropriate ethical standards in conducting research and writing the manuscript, following all applicable laws and regulations regarding treatment of animals or human subjects.}

\coi{We declare we don't have conflicts of interest.}

\data{We trained on data available from ABCD and HCP-D studies, available online through the NIH portals. We provide scripts for generating the train/validation/test splits that were used for training our models.}

\bibliography{main}

\end{document}